\documentclass[letterpaper, 10 pt, conference]{ieeeconf}  

\IEEEoverridecommandlockouts                              

\usepackage[ruled,norelsize]{algorithm2e}
\usepackage{graphicx}
\usepackage{subcaption}
\usepackage{hyperref}

\graphicspath{{}}

\makeatletter
\newcommand{\removelatexerror}{\let\@latex@error\@gobble}
\makeatother

\title{\LARGE \bf
AA-ICP: Iterative Closest Point with Anderson Acceleration
}

\author{A.L. Pavlov$^{1}$, G.V. Ovchinnikov$^{2}$, D. Yu. Derbyshev$^{3}$, D. Tsetserukou$^{4}$ and I.V. Oseledets$^{5}$
\thanks{*The research was supported by RSF (project No. 17-11-01376)}
\thanks{$^{1}$A.L. Pavlov is a PhD student with Space Center,
        Skolkovo Institute of Science and Technology, Skolkovo Innovation Center, Moscow 143026 Russia
        {\tt\small artem.pavlov@skolkovotech.ru}}%
\thanks{$^{2}$G.V. Ovchinnikov is a Research Scientist with the Center for Computational and Data-Intensive Science and Engineering,
        Skolkovo Institute of Science and Technology
        {\tt\small ovgeorge@yandex.ru}}%
\thanks{$^{3}$D. Yu. Derbyshev is a PhD student with Moscow Institute of Physics and Technology}%
\thanks{$^{4}$D. Tsetserukou is a Associate Professor with Space Center,
        Skolkovo Institute of Science and Technology
    {\tt\small D.Tsetserukou@skoltech.ru}}%
\thanks{$^{4}$I.V. Oseledets is a Associate Professor with the Center for Computational and Data-Intensive Science and Engineering,
        Skolkovo Institute of Science and Technology
    {\tt\small I.Oseledets@skoltech.ru}}%
}

\begin{document}

\maketitle
\thispagestyle{empty}
\pagestyle{empty}

\begin{abstract}

Iterative Closest Point (ICP) is a widely used method for performing scan-matching and registration.
Being simple and robust method, it is still computationally expensive and may be 
challenging to use in real-time applications with limited resources on mobile platforms.
In this paper we propose novel effective method for acceleration of ICP which does not require 
substantial modifications to the existing code. 
        
This method is based on an idea of Anderson acceleration which is an iterative procedure for finding a fixed point of contractive mapping. The latter is often faster than a standard Picard iteration, usually used in ICP implementations. We show that ICP, being a fixed point problem, can be significantly accelerated by this method enhanced by heuristics to improve overall robustness. We implement proposed approach into Point Cloud Library (PCL) and make it available online. Benchmarking on real-world data fully supports our claims.

\end{abstract}

\section{Introduction}

The localization is one of the fundamental problems of modern robotics. 
Robots commonly use the laser scanner data presented in the form of point clouds. 
To relate one scan to another scan-matching algorithms are used with Iterative Closest Point  \cite{besl1992method} being one of the most popular approaches. Scan-matching problems also 
arise outside of the robotics domain, for example, in the context of 
medical image comparison and registration \cite{stewart2003dual}, \cite{ge1996surface}, 
for which ICP is widely used as well. 

In practice, different modifications are implemented to speed up the matching process and to improve the ICP reliability. (see section \ref{sec:icp})
Nevertheless, underlying iterative structure of ICP is general remains almost unchanged. 
In this paper, we propose to accelerate ICP through modification of iteration procedure, so we can keep 
all benefits from the above-mentioned modifications, thus making ICP even faster. Instead of using "state-less" approach which depends only on the last iteration, the proposed idea is to select the next iteration point based on solution of specific optimization problem over history of previous iterations. 
The optimization problem itself is quite simple and therefore can be solved on any hardware, 
which is capable of running ICP in the first place.

This paper is organized as follows. In section 2 we provide theoretical background by describing basic form of ICP and Anderson acceleration, and then
highlight the main theoretic properties for both.
Next, we propose modifications necessary to make Anderson accelerated version of ICP (AA-ICP) to work with the real world data, which is often violates underlying assumptions. In section 3 we provide experimental results, which prove that our modification of ICP with Anderson acceleration achieves significant speed-up (more than $30\%$) and the slight error improvement ($0.3\%$) on dataset from \cite{sturm2012benchmark}.

In this work we focus on 3D datasets, but proposed method can be applied to 2D ICP variants as well.

We implement AA-ICP as part of the widely used Point Cloud Library (PCL), and the source code is freely available in our fork repository\footnote{\url{https://github.com/SkoltechRobotics/pcl/tree/anderson}}. Finally, section 4 summarizes the paper.

\begin{figure}
    \begin{subfigure}[b]{0.49\columnwidth}
        \includegraphics[width=\textwidth]{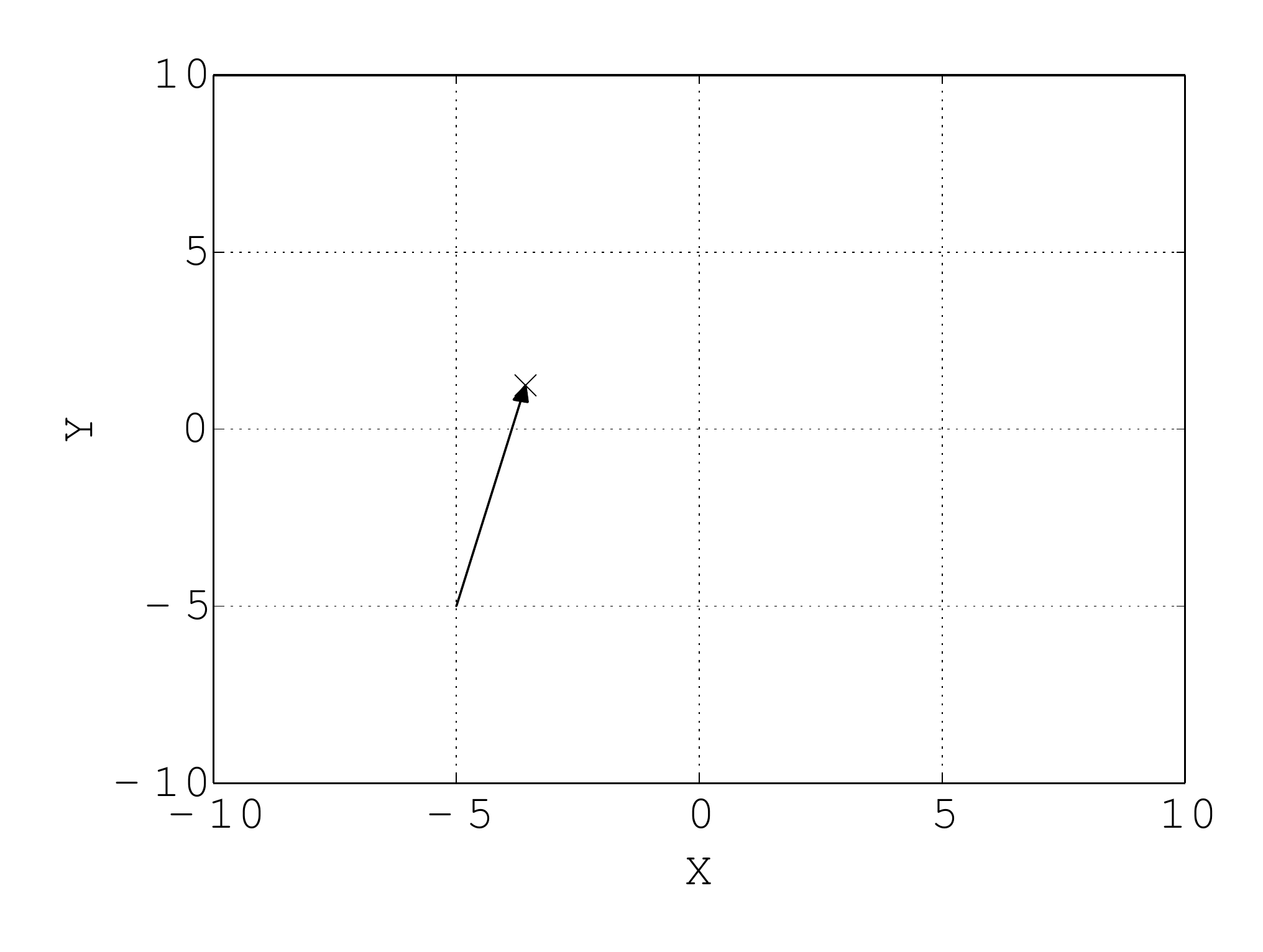}
        \caption{$n=1$}
    \end{subfigure}
    \begin{subfigure}[b]{0.49\columnwidth}
        \includegraphics[width=\textwidth]{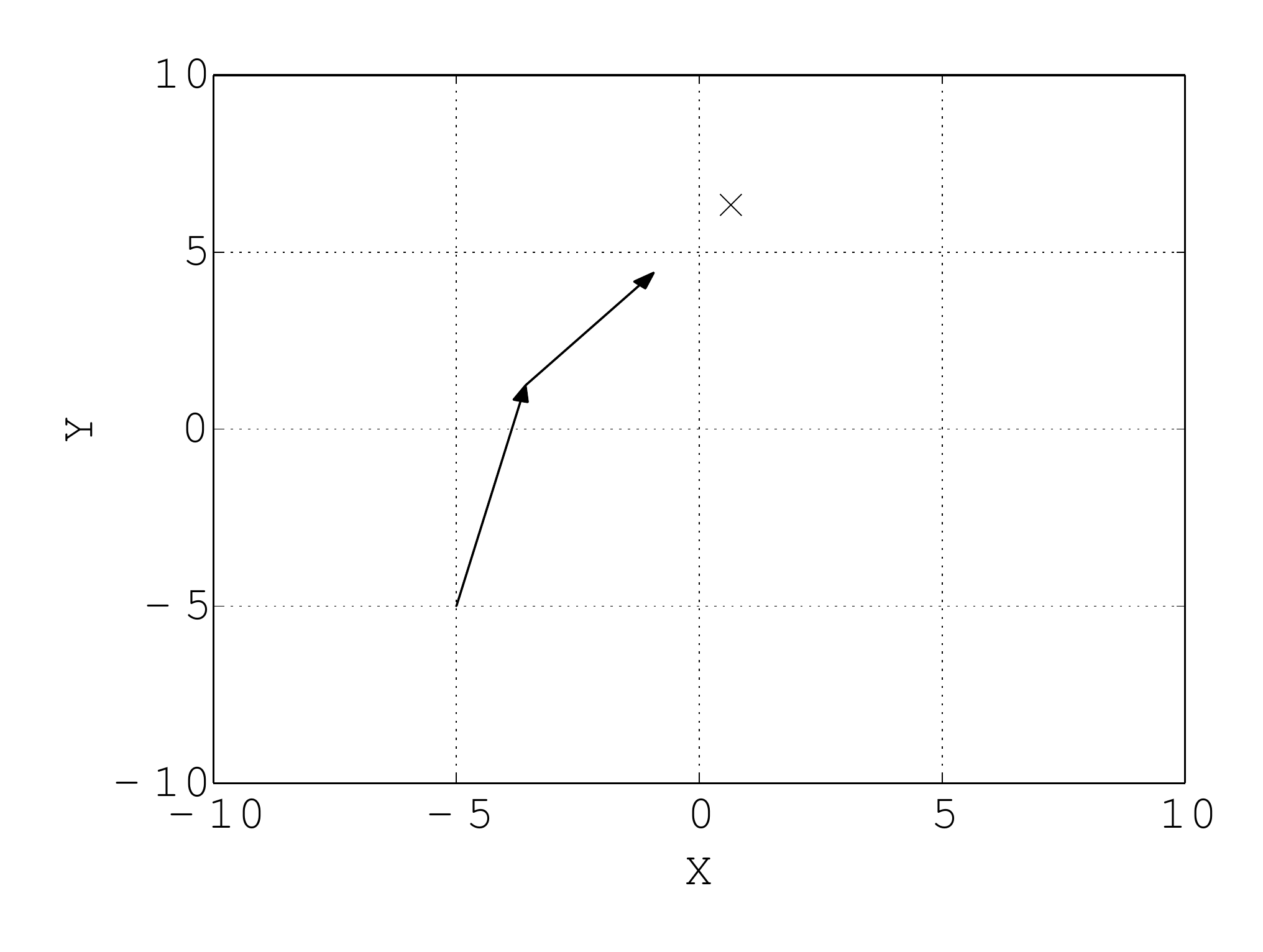}
        \caption{$n=2$}
    \end{subfigure}
    \begin{subfigure}[b]{0.49\columnwidth}
        \includegraphics[width=\textwidth]{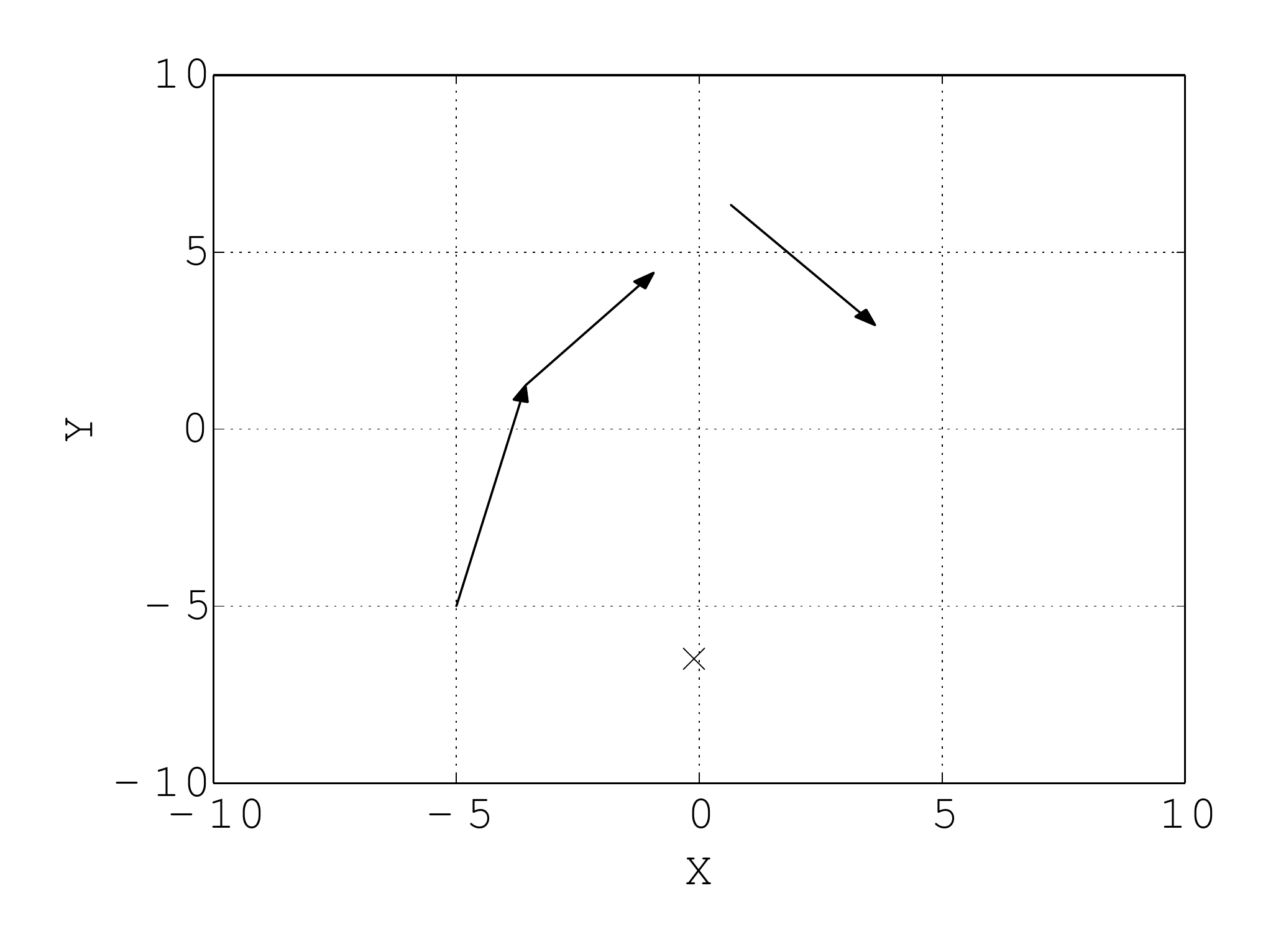}
        \caption{$n=3$}
    \end{subfigure}
    \begin{subfigure}[b]{0.49\columnwidth}
        \includegraphics[width=\textwidth]{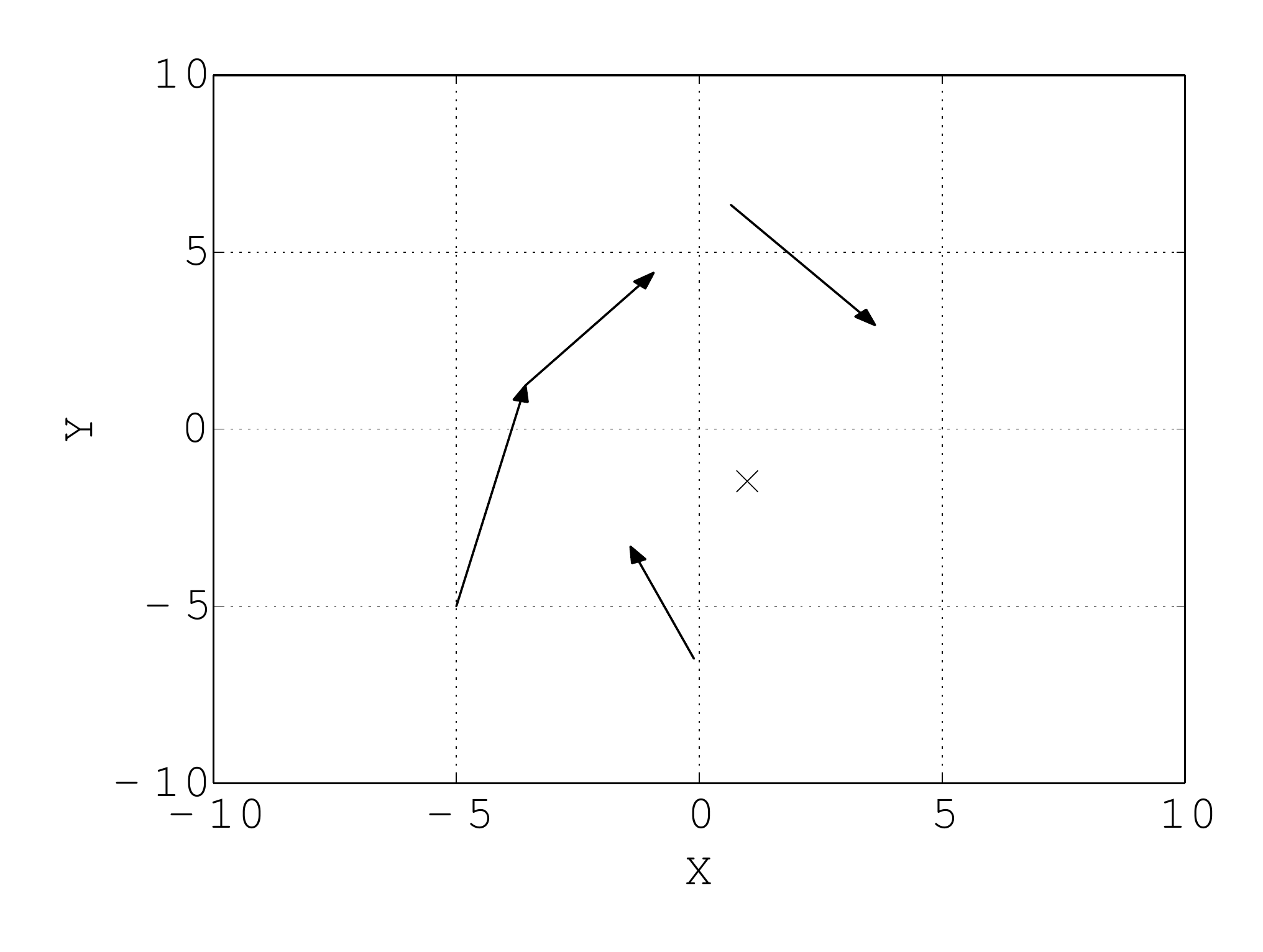}
        \caption{$n=4$}
    \end{subfigure}
    \caption{Demonstration of Anderson acceleration behaviour for simple 2D case with whirlpool-like mapping; $n$ is the number of iteration, vectors denote mapping $G(u)$, cross mark shows position of $u_{n+1}$}
    \label{fig:anderson_demo}
\end{figure}

\section{Background}
There are many variations of ICP: \cite{chetverikov2002trimmed}, \cite{censi2008icp} and \cite{men2011color} available nowadays, but for our discussion it is important that ICP essentially boils down to a fixed point problem:
\begin{equation}
\label{fixedpoint}
    u = G(u),
\end{equation}
which is usually solved with the simple (or also called Picard) iteration procedure: $u^{k+1} = G(u^k)$. Here and thereafter $u \in \mathbf{R}^{n}$ and describes the roto-translation between two scans ($u$ can be represented by translation coordinates and Euler angles for rotation).

Anderson acceleration \cite{anderson1965iterative} 
(also known as Anderson mixing in computational chemistry)
is a different method of finding $u^{k+1}$ based on the history of $m+1$ latest 
iterations and residuals. In case of $m=0$ it is equivalent to Picard iteration.

In general, Anderson acceleration is theoretically and practically superior to Picard in many cases \cite{walker2011anderson}, \cite{toth2015convergence} and additional cost of selecting different points is equal to solving least squares problem of size $m$, which for ICP is negligible compared to the cost of a single iteration. 

\subsection{Iterative Closest Point}
\label{sec:icp}
ICP algorithms are often used in robot navigation for performing scan-matching of data provided by LIDARs, RGB-D cameras or stereo-cameras. 
Thus, it is quite common underlying algorithm in different simultaneous localization and mapping (SLAM) setups as discussed in \cite{frese2006discussion} and in a two-part survey \cite{durrant2006simultaneous}, \cite{bailey2006simultaneous}.

ICP operates on two sets of points $S$ and $S^{ref}$, where $S_i, S^{ref}_j \in \mathbf{R}^{n}$ (usually $n=2$ or $n=3$) , with initial proper rigid transformation guess $u^0$. The simplest algorithm variant \cite{besl1992method} can be described as follows:
\begin{enumerate}
\item Transform $S$ using $u^0$.
\item For every point in $S$ find the closest point in $S^{ref}$, pairs of such points are called correspondencies.
\item Find such transform $u$ which minimizes the mean distance between correspondencies (i.e. error).
\item Apply transformation $u$ to $S$.
\item If change in the error falls below a given threshold - then terminate; otherwise - go to step 2.
\end{enumerate}

As mentioned before various modifications are usually used in practice, such as:
\begin{itemize}
	\item Different metrics (point-to-plane, feature based)
	\item Usage of indexes (e.g. K-d trees)
	\item Dynamic caching
	\item Point sampling
	\item Random restarts
	\item And others
\end{itemize}

But in general those modification still use basic Picard iteration.

\subsection{Anderson acceleration}

We notice several properties of ICP, which make finding fixed point of eq. \ref{fixedpoint}
computationally expensive and hard to accelerate with higher-order methods.

The main problem is the unavailability of derivatives because function $G$ in ICP case is not even differentiable. While this limitation can be side-stepped with continuous relaxation (see, for example \cite{biber2006nscan} applying Newton-Raphson method to ICP formulated in terms of differentiable energy functions), even with continuous relaxation computation of gradients is often prohibitively expensive. Moreover, continuous relaxation is both hard to implement from scratch and to apply to the existing code, e.g., to PCL\footnote{\url{http://pointclouds.org/}} or CSM\footnote{\url{ttps://censi.science/software/csm/}}, which
are widely used in community. The other option could be the 
usage of numerical differentiation, 
but its computational cost is also high. Additionally, due to noise in the
input data finite-difference
approximation of gradients becomes unreliable.

As was outlined above, in this work we propose to use Anderson acceleration to speed up the convergence process of ICP. For
differentiable functions it is analogous  to pseudo-Newton methods \cite{fang2009two}
and in linear case it is equivalent to GMRES \cite{potra2013characterization}.

In our perspective, Anderson acceleration is the best iteration scheme we could 
employ here: it almost always requires less iterations to converge to the same error
than simple iteration; it does not require additional calls to the expensive and memory demanding ICP function;
it relies on the history of iterations alone with overhead being negligible
compared
to the cost of a single ICP iteration step. Last, but not least, it could be trivially
added into 
the existing ICP implementations, 
thus many existing implementations of ICP could benefit from it.

The simple variant of Anderson acceleration is represented by algorithm \ref{alg:simple_aa}.

\begin{algorithm}
    \KwData{initial guess $u^0$, contraction mapping $G$, maximum iterations limit $n_{max}$}
    \KwResult{fixed point $u$}
    $g^0 = G(u^0)$\;
    $f^0 = g^0 - u^0$\;
    $u^1 = g^0$\;
    \For{$n$ in $1..n_{max}$}{
      $f^n = G(u^{n}) - u^{n}$\;
      Find $\alpha \in \mathbf{R}^{n+1}$ which minimizes
        $\|\sum_{j=0}^n \alpha_j f^j \|_2$ subject to $\sum_j \alpha_j = 1$\;
      $u^{n+1} = \sum_{j=0}^{n} \alpha_j G(u^j)$\;
      \If{convergence criteria is true}{
        \textbf{break};
      }
   }
   \Return $u^n$\;
 \caption{Anderson acceleration}
 \label{alg:simple_aa}
\end{algorithm}

Here $a..b$ denotes sequence of integers from $a$ (including) to $b$ (excluding), so $1..4$ implies the following sequence: 1, 2 and 3.

The minimization problem for $\alpha$ can be reformulated as the linear least-squares problem.
Constraint can be replaced by substitution $\alpha_0 = 1 - \sum_{j=1}^{n} \alpha_j$, which leads to the following unconstrained problem:
$$
\min_{\alpha_1, \dots, \alpha_n} \| f_0 + \sum_{i=1}^n \alpha_i (f_i - f_0) \|.
$$

The behavior of AA for 2D translations (without rotation) can be is presented in fig. \ref{fig:anderson_demo}. It clearly showcases ``jumpy'' nature of the algorithm, which enables much faster convergence compared to simple Picard iteration. Such jumps can be viewed as attempts to predict the most plausible convergence point based on the history of previous iterations.

\subsection{Heuristics}

The main problem of Anderson acceleration is the serious instability when working with non-contractive mappings, which is quite common when processing real-world data. 
For example, existence of several convergence points 
automatically makes related mapping non-contractive. 
Even the existence of a single convergence point does not make mapping contractive.
To illustrate this, imagine a drainage basin, in which all water flows towards single lake, but on the ridge, which divides two tributaries, water will flow in different directions.

Due to this instability, in practice we cannot use simple AA demonstrated in alg. \ref{alg:simple_aa}. Usually various heuristics and modifications can be added, such as: limiting history length; introduction of dumping factors; linear search using easier to compute but less precise approximations. 
In this work we propose a specific set of heuristics developed considering ICP properties.

Our final algorithm can be represented by alg. \ref{alg:aa_heuristic}.

\begin{algorithm}
    \KwData{initial guess $u^0$, history length limit $m$, alpha limit $\alpha_l$, maximum iterations limit $n_{max}$}
    \KwResult{convergence point $u^{n+1}$}
    $h = 0$\tcp*{history cut-off cursor}
    $g^0 = G(u^0)$\;
    $f^0 = g^0 - u^0$\;
    $u^1 = g^0$\;

    \For{$n$ in $1..n_{max}$}{
    $g^n = G(u^{n})$\;
    $f^n = g^n - u^{n}$\;
    \If{ICP error is too big}{
      $h = n$\tcp*{"resetting" history}
      $u^{n+1} = g^{n-1}$\;
      \textbf{continue};
    }
    $u^{n+1} = g^n$\;
    \For{$i$ in $1..min(m, n-h)$}{
        $\alpha_{1, \dots, n} = \min \| f_0 + \sum_{j=1}^n \alpha_j (f_j - f_0) \|$\;
        $\alpha_0 = 1 - \sum_{j=1}^{n} \alpha_j$\;
      \uIf{$(-\alpha_l \le \alpha_j \le \alpha_l, \forall j)$ and $(\alpha_0 > 0)$ }{
        $u^{n+1} = \sum_{j=0}^{i} \alpha_j g^{n-j}$\;
      }
      \Else {
        \textbf{break};
      }
    }
    
      \If{convergence criteria is true}{
        \textbf{break};
      }
 
    }
    \Return $u^{n+1}$\;
 \caption{AA-ICP}
 \label{alg:aa_heuristic}
\end{algorithm}

In this algorithm we use $u$ for denoting concatenated vector of translation coordinates and Euler angles for rotation, e.g. $u = (x, y, z, \phi, \theta, \psi)$ , thus for 3D case $u \in \mathbf{R}^6$ and for 2D case $u \in \mathbf{R}^3$. 
Due to assumption that two given scans are spatially close, singularity in Euler representation of rotation does not impact us. It could be argued that reduction of roto-translation for 3D case to such vector and operating with it as described here is not strictly correct, since the addition of rotations is not a commutative operation.

The first heuristic in algorithm checks whether the error estimate returned by ICP step is not 
considerably large than error for previous iteration 
(the most common error estimate is the mean distance between correspondences).
In case this condition is true, the history is reset iterations start from $g^{n-1}$,
i.e. a full reset of history occurs and we leave only last known ``trustworthy'' point. 
It's a safeguard against cases, when the second heuristic fails to filter out bad jumps. 
This heuristic results in ``empty'' iteration, i.e. only error estimate is used from ICP step call, but such cases are relatively rare ($3-5\%$ of iterations) and associated costs are out-weighted by provided robustness.

The second heuristic loops over the incremented history length, which is used for Anderson acceleration and 
checks whether computed alphas fall within the specified range. 
It also checks whether $\alpha_0 > 0$, thus ensuring that jump occurs in general direction defined by last $G$ call. 
In case those conditions are not met, the last result is returned. 
This way only points most probably residing inside a local contractive area are selected for the calculations. It also ensures that the next jump doesn’t occur in the reverse direction in respect to the last ICP step.

The overall cost of solving several small linear systems is negligible compared to the cost of $G$, as in practice $i$ in the alg. \ref{alg:aa_heuristic} rarely exceeds 5-10.

\section{Experimental Results}

To measure the performance of AA-ICP we modified Point Cloud Library (PCL) and added AA-ICP implementation. The modified code is freely available in our fork repository\footnote{\url{https://github.com/SkoltechRobotics/pcl/tree/anderson}} and it is intended for the inclusion into the upstream. 
Tooling used for acquiring data are also available in the separate repository. \footnote{\url{https://github.com/SkoltechRobotics/aa-icp-tools}}

We used two datasets for our experiments: ``RGB-D SLAM Dataset and Benchmark'' \cite{sturm2012benchmark} and the ``Stanford Bunny'' \cite{turk2005stanford}. 
The first one was collected using RGB-D camera and the second -- with the laser range scanner.

\subsection{RGB-D SLAM Dataset and Benchmark}

The following sequences from freiburg1 set was used: room, desk and xyz. To emulate keyframes we matched not subsequent frames, but the 5th scan from the current one. Thus, in total 2738 scan pairs were processed without any filtering or sampling.

\begin{figure}
    \begin{subfigure}[b]{0.49\columnwidth}
        \includegraphics[width=\textwidth]{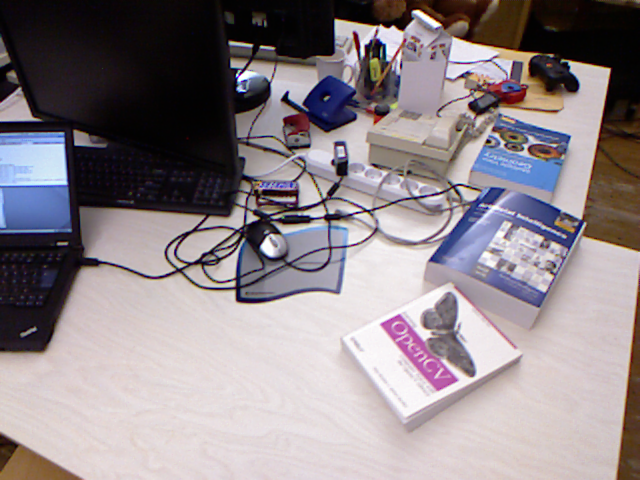}
    \end{subfigure}
    \begin{subfigure}[b]{0.49\columnwidth}
        \includegraphics[width=\textwidth]{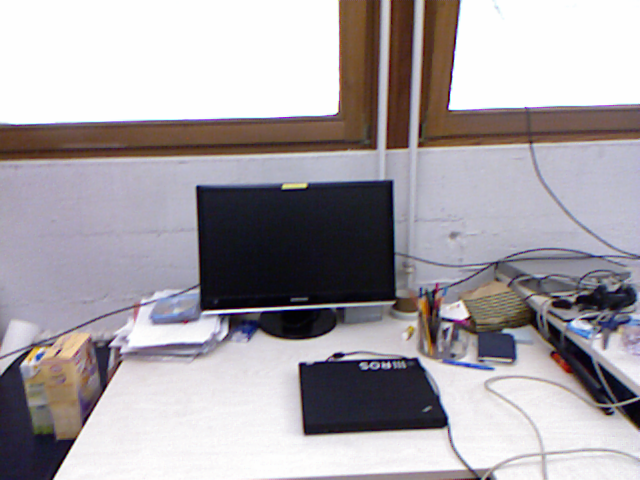}
    \end{subfigure}
    \caption{Frame examples from Freiburg dataset}
    \label{fig:freiburg_frames}
\end{figure}

The examples of RGB images are presented in fig. \ref{fig:freiburg_frames}. In our experiments only depth channel was used.

\begin{figure}
    \includegraphics[width=\columnwidth]{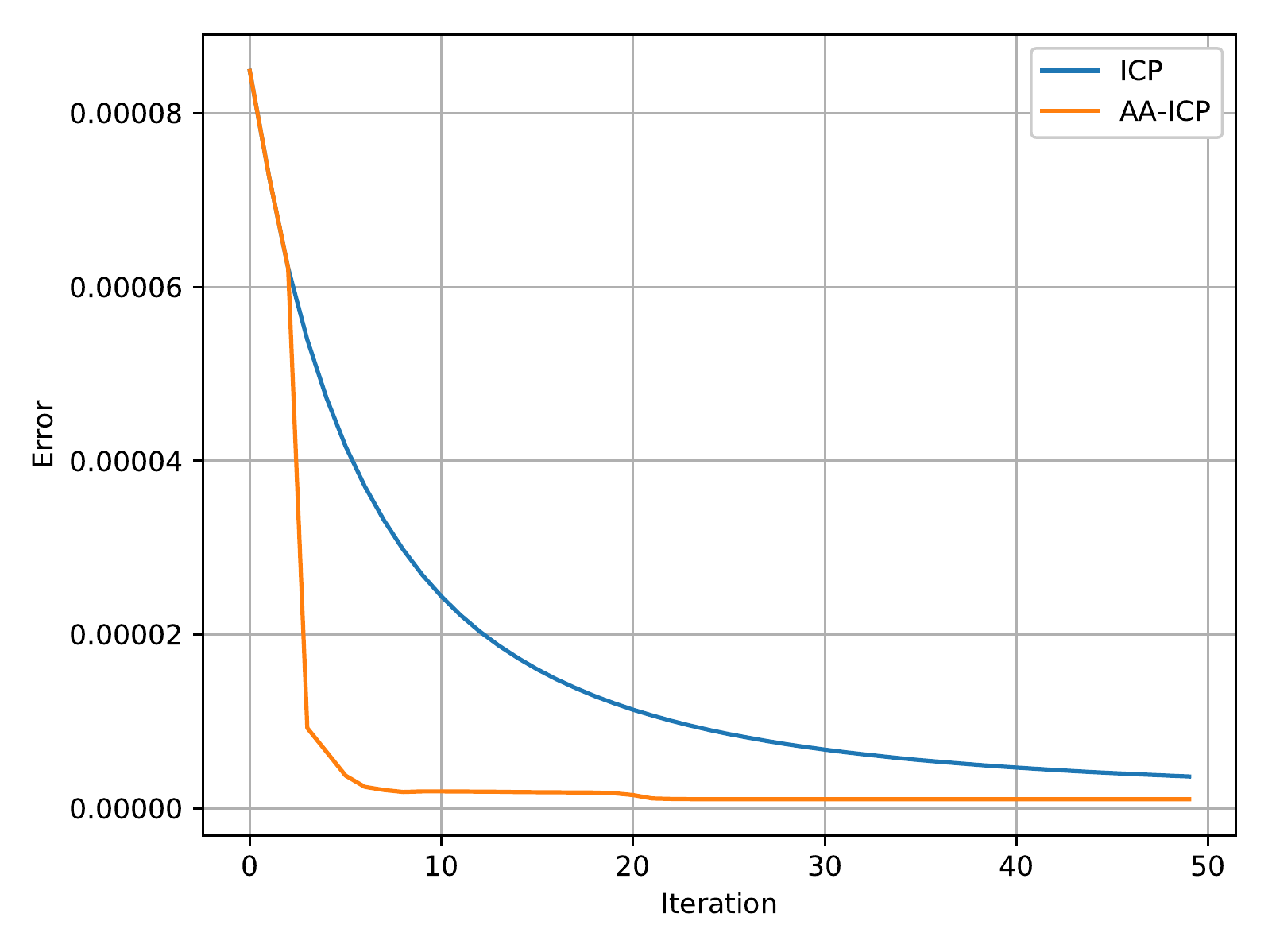}
    \caption{Example of error estimate behaviour for simple ICP and AA-ICP}
    \label{fig:errors_comp}
\end{figure}

Example of error behaviours for AA-ICP and default PCL ICP is shown in fig.\ref{fig:errors_comp} . Both methods generally converge to the same point.

However these criteria in approximately $1\%$ of cases terminated AA-ICP too early, 
thus algorithm was improved by requiring the convergence criteria to hold true for two iterations in a row. It addeds one iteration to AA-ICP (although with smarter convergence criteria it can be removed). Nevertheless, even with such handicap statistically AA-ICP converged faster than simple ICP. There was one exception though: if criteria were satisfied earlier than 4th iteration, the confirming iteration was not required, since this indicated that camera movement between scans was negligible. The maximum number of iterations was limited to 100.

\begin{figure}
    \includegraphics[width=\columnwidth]{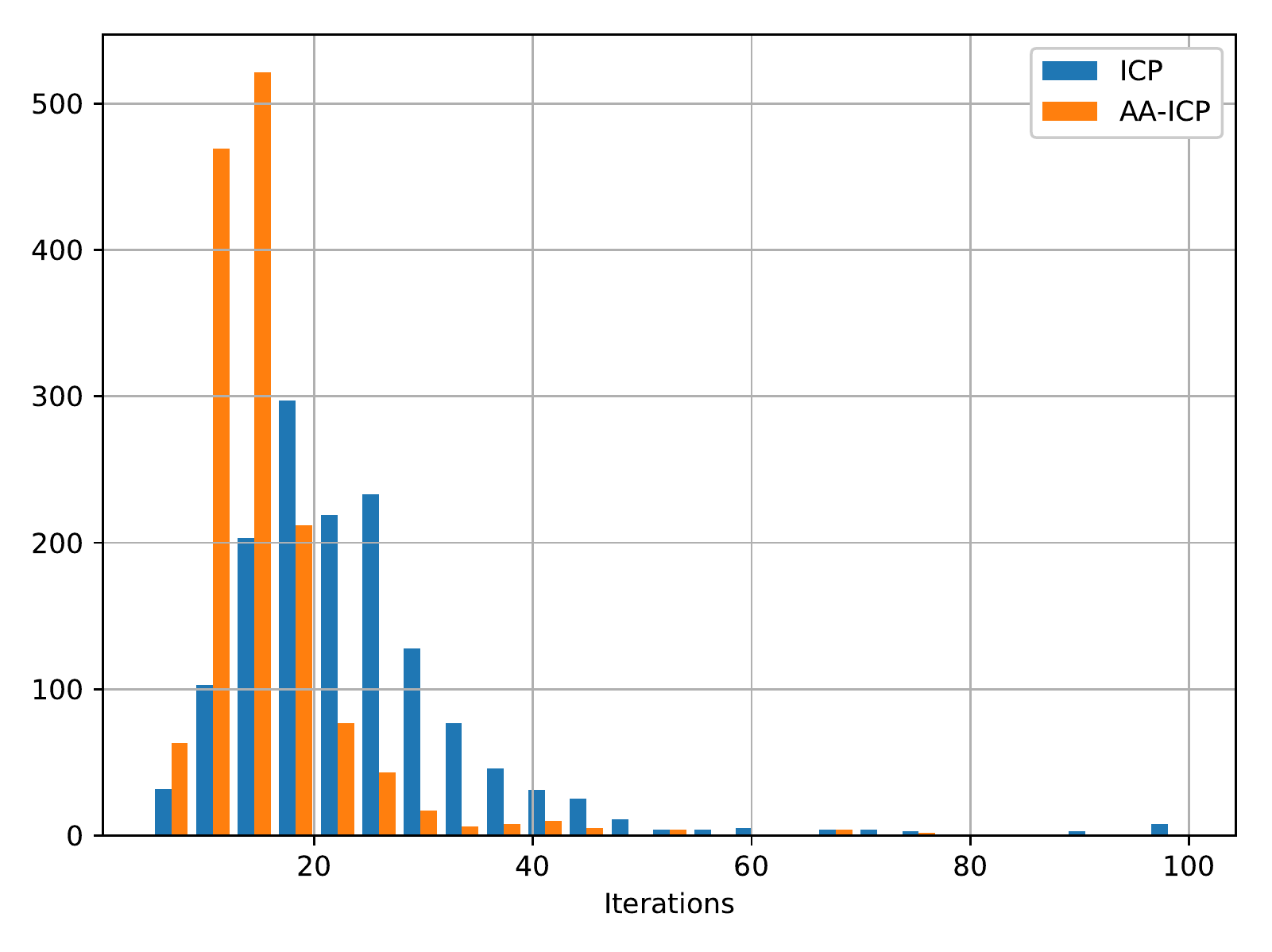}
    \caption{The iterations number required for the simple ICP and AA-ICP to converge}
    \label{fig:abs_iters}
\end{figure}

\begin{figure}
    \includegraphics[width=\columnwidth]{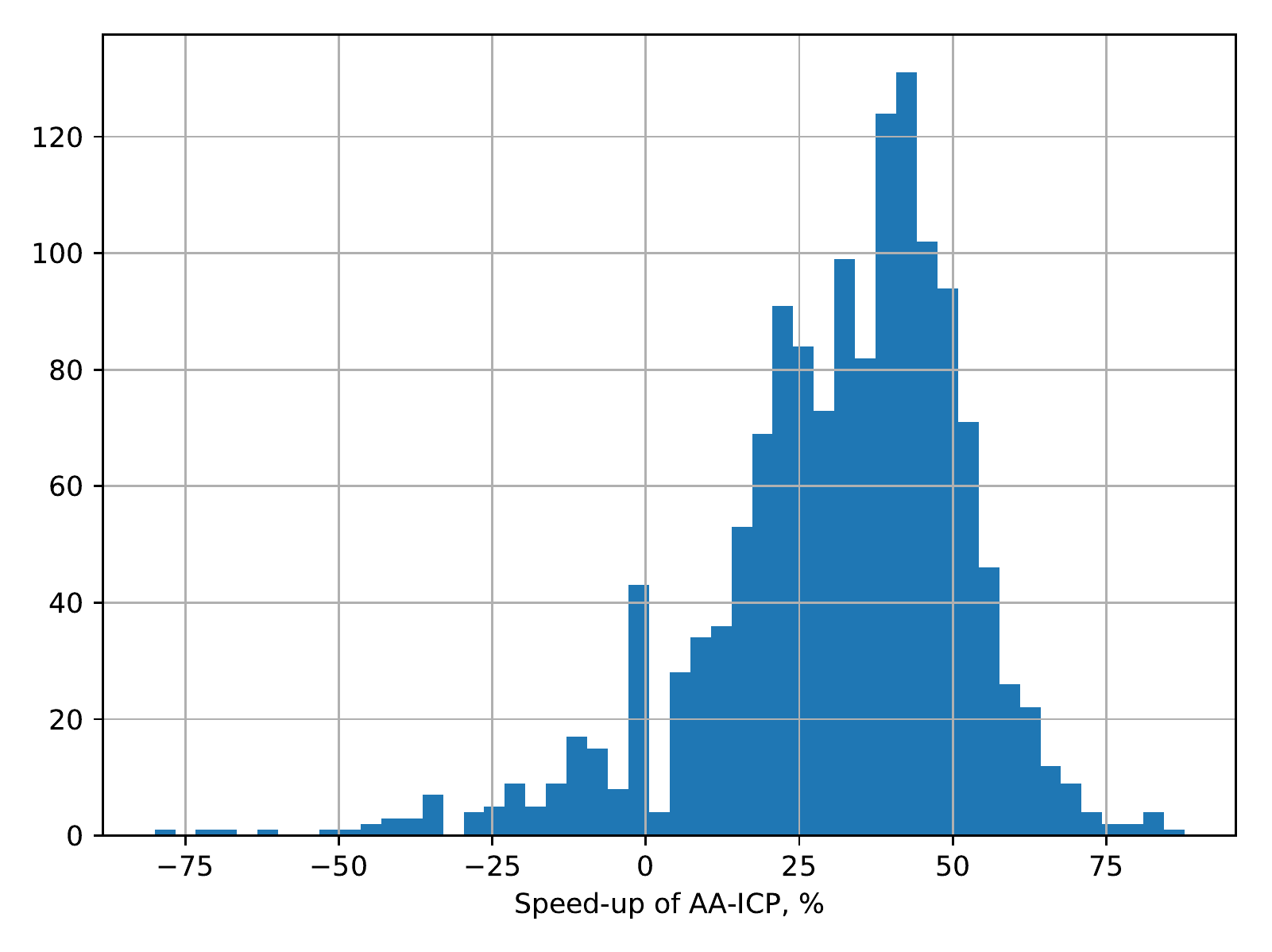}
    \caption{AA-ICP speed-up relative to the simple ICP}
    \label{fig:rel_iters}
\end{figure}

Fig. \ref{fig:abs_iters} demonstrates the statistical properties of acceleration over the number of iterations required for convergence, which were calculated for $\varepsilon = 0.001$ \footnote{See documentation for \texttt{setEuclideanFitnessEpsilon} method of \texttt{pcl::Registration} class} and $\alpha_l = 10$. Fig. \ref{fig:rel_iters} shows the relative change between the number of iterations for the same pair of scans. (see fig.) Median speed-up in this case equals to 35\%, mean to 30\%, and overall more than 90\% of AA-ICP runs got accelerated relative to the simple ICP.

\begin{figure}
    \includegraphics[width=\columnwidth]{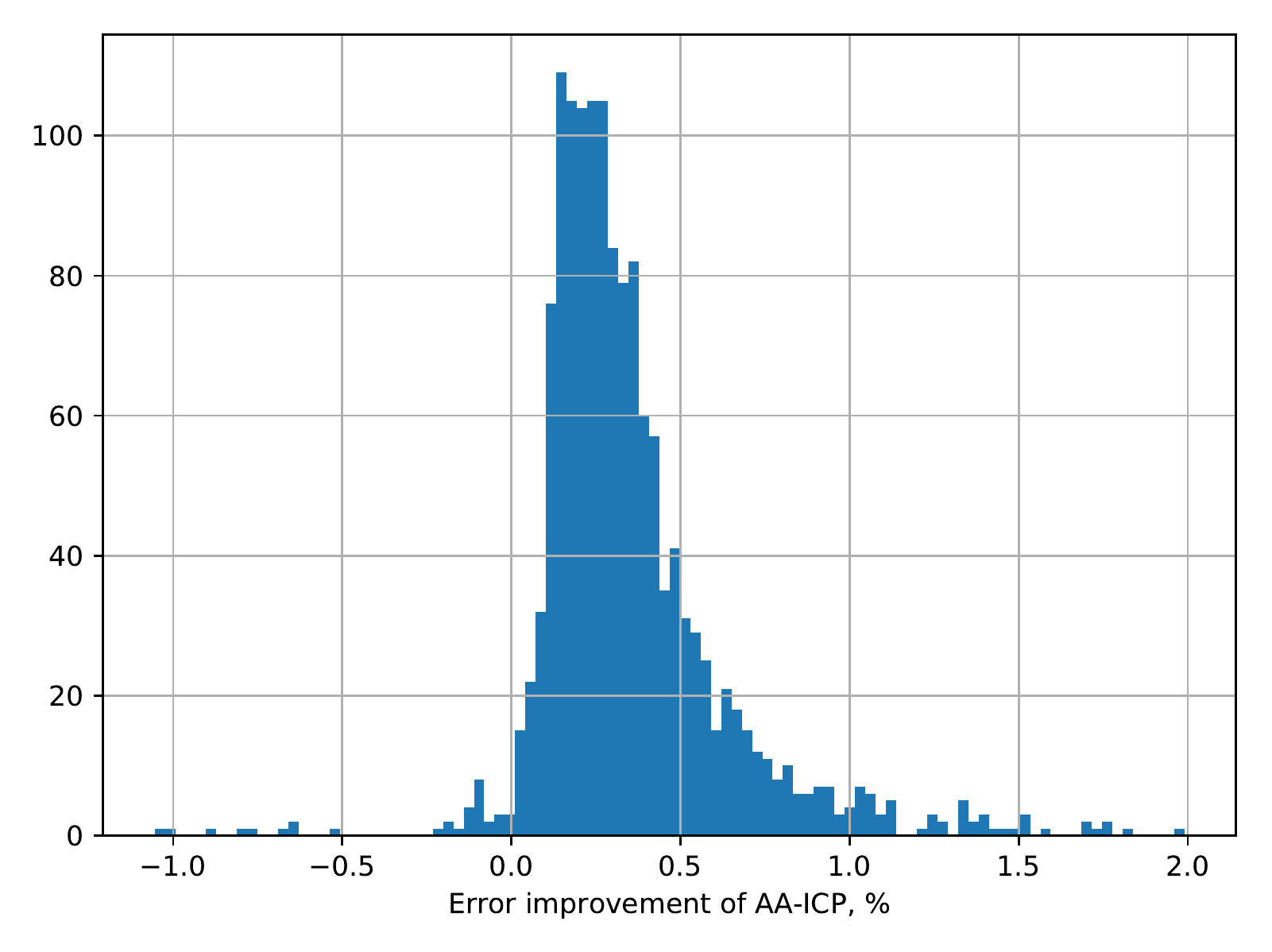}
    \caption{Final error improvement of AA-ICP relative to the simple ICP}
    \label{fig:rel_errs}
\end{figure}

In addition to the smaller number iterations required for convergence, AA-ICP generates results with better quality. This can be seen from fig. \ref{fig:rel_errs}, which depicts relative final errors produced by AA-ICP and the simple ICP. More than 97\% of runs converged to smaller errors with AA-ICP. The median improvement equals to approximately 0.3\% and mean to 0.4\%. Note that convergence criteria modification does not provide a big contribution to those values, as a single iteration of nearly converged AA-ICP usually results in error change with order of magnitude equal of $0.001\%-0.01\%$.

\begin{figure}
    \includegraphics[width=\columnwidth]{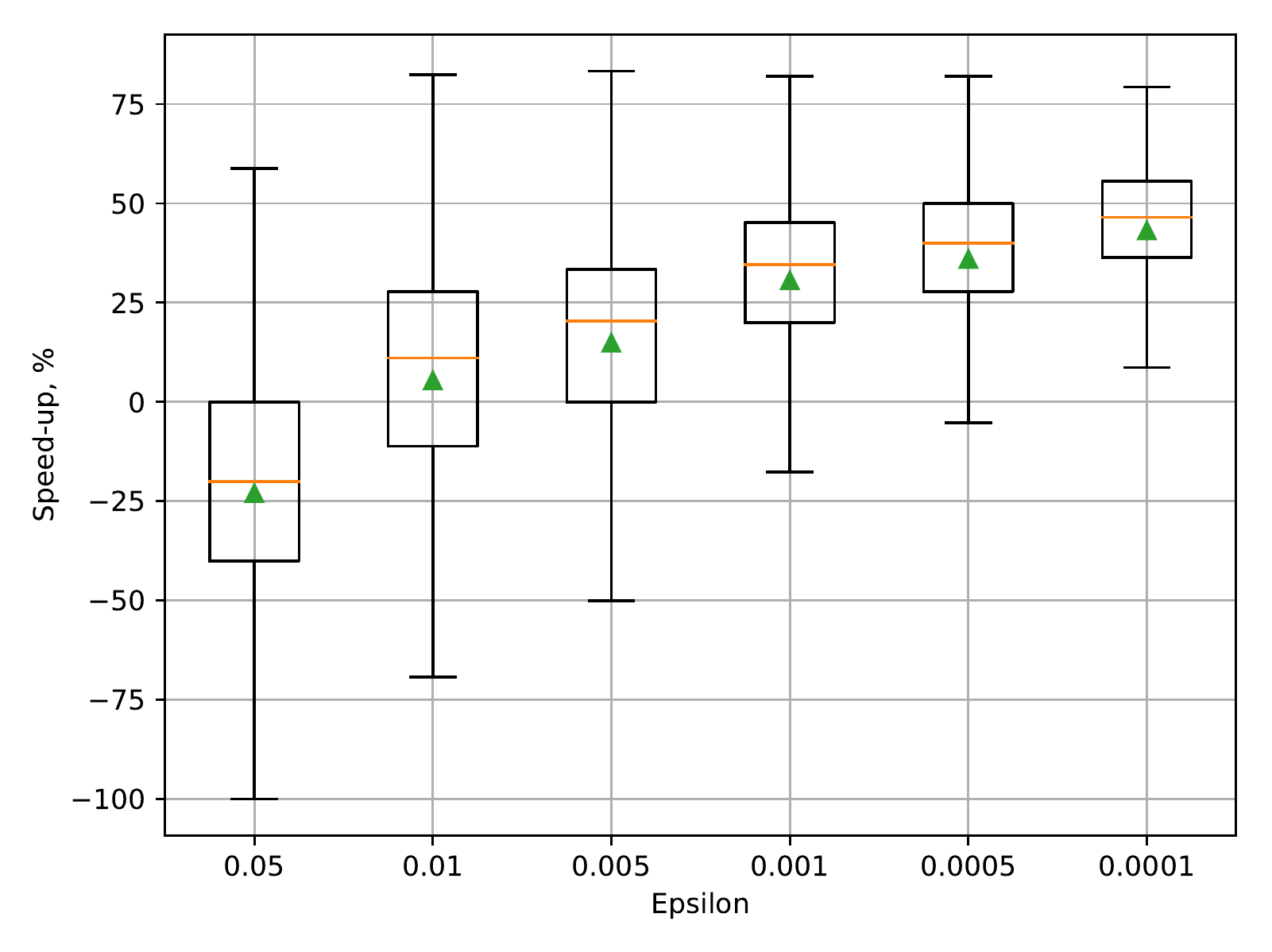}
    \caption{Box-plot of relative AA-ICP speed-up depending on $\varepsilon$ value for RGB-D dataset. Triangle mark denotes mean value.}
    \label{fig:eps_dep}
\end{figure}

Finally we demonstrate dependence between $\varepsilon$ and acceleration improvement in fig. \ref{fig:eps_dep}. As we can see relative speed-up of AA-ICP increases with rising quality criteria for final results, while if convergence criteria is overly-relaxed AA-ICP results in bigger number of iterations.

\subsection{Stanford Bunny}

\begin{figure}
	\centering
    \includegraphics[width=0.7\columnwidth]{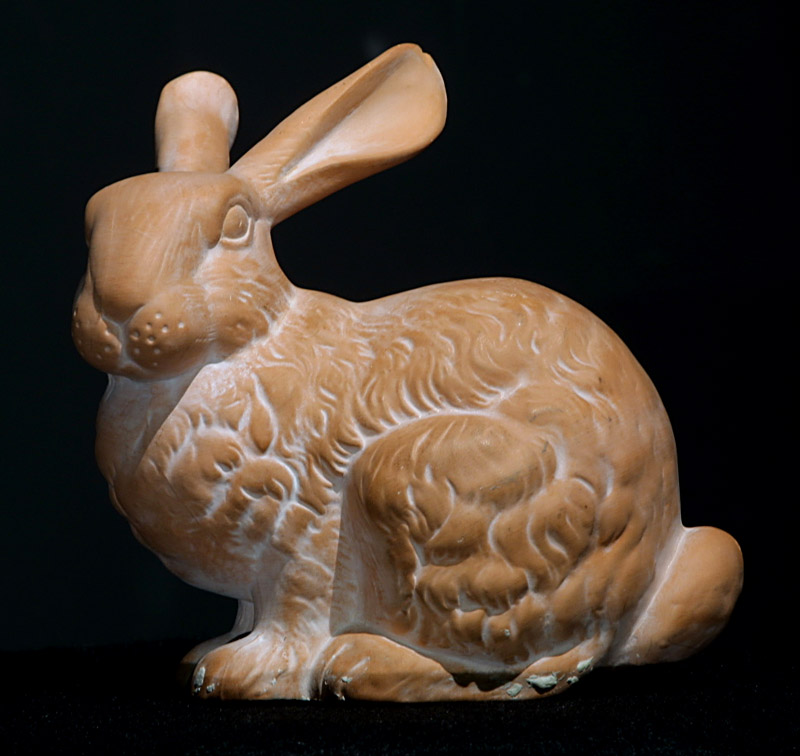}
    \caption{Photo of "Stanford bunny"}
    \label{fig:bunny}
\end{figure}

\begin{figure}
    \begin{subfigure}[b]{0.49\columnwidth}
        \includegraphics[width=\textwidth]{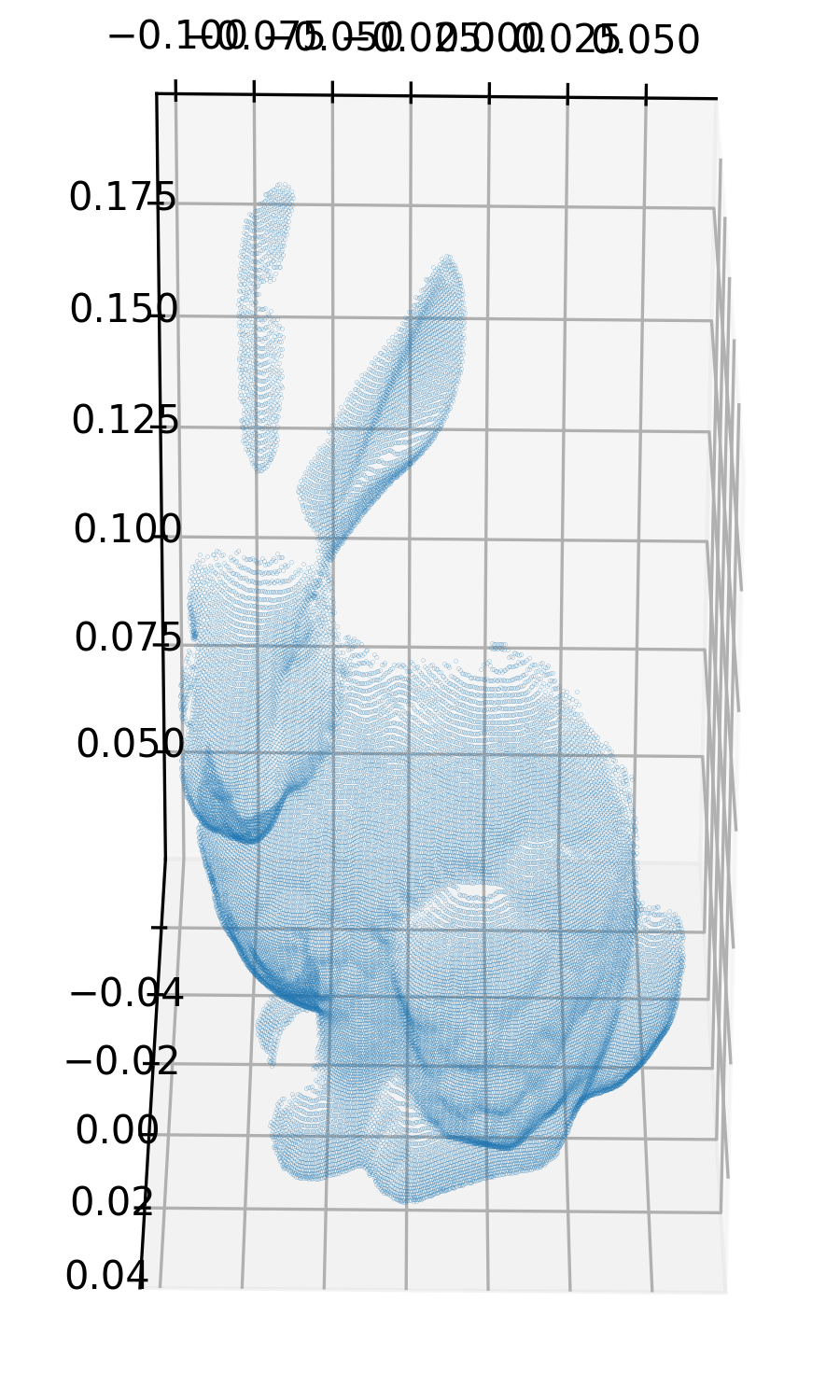}
        \caption{bun000}
    \end{subfigure}
    \begin{subfigure}[b]{0.49\columnwidth}
        \includegraphics[width=\textwidth]{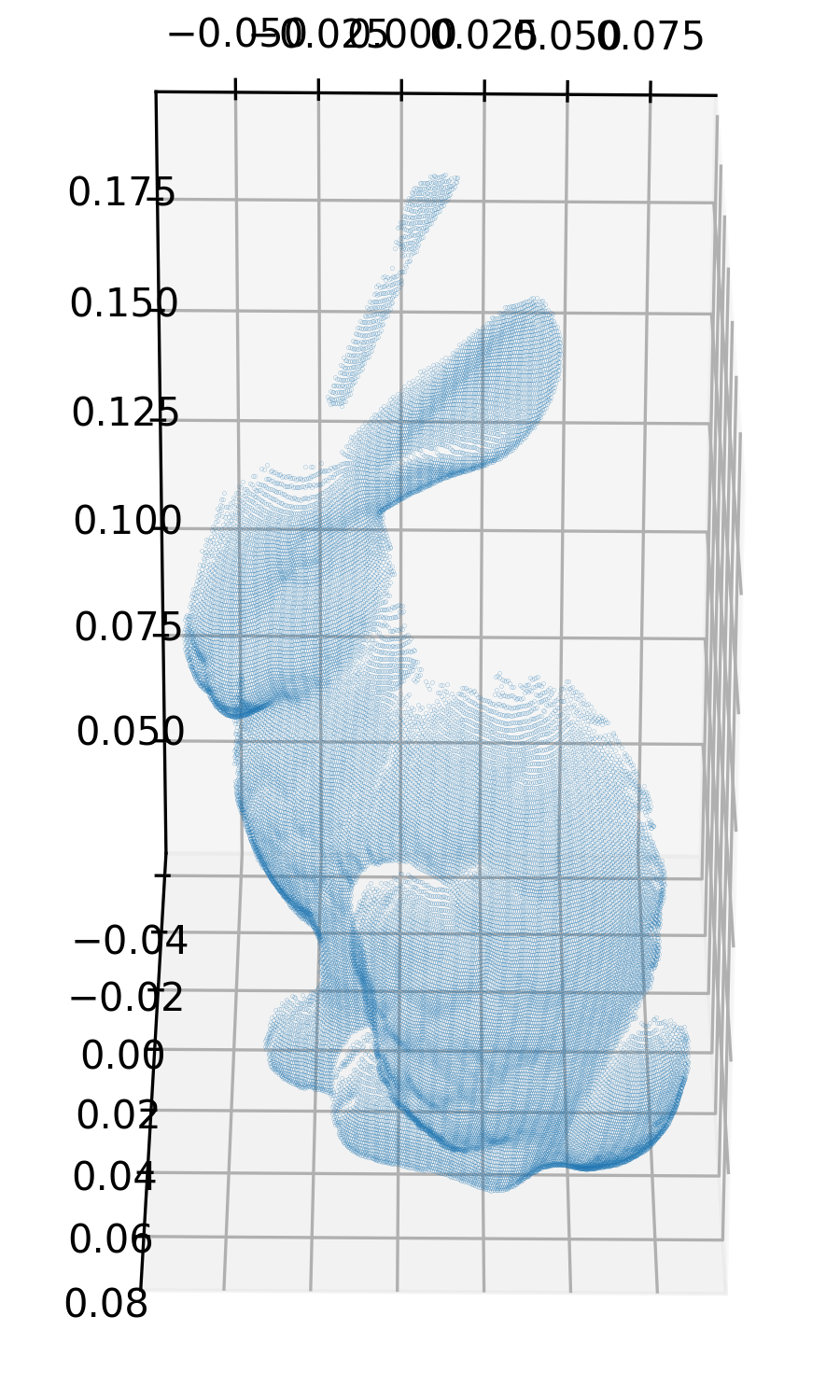}
        \caption{bun045}
    \end{subfigure}
    \caption{Point clouds of "Stanford bunny" used in our work}
    \label{fig:bunny_scans}
\end{figure}

To test acceleration properties of AA-ICP depending on misalignment of scans we chose to use the "Standford Bunny" \footnote{\url{http://graphics.stanford.edu/data/3Dscanrep/}}, probably one of the most well known 3D test models taken by laser scanner. You can see its photo in fig. \ref{fig:bunny} and point clouds in fig. \ref{fig:bunny_scans}. It provides highest quality ground truth data for performing initial scan alignment.

\begin{figure}
    \includegraphics[width=\columnwidth]{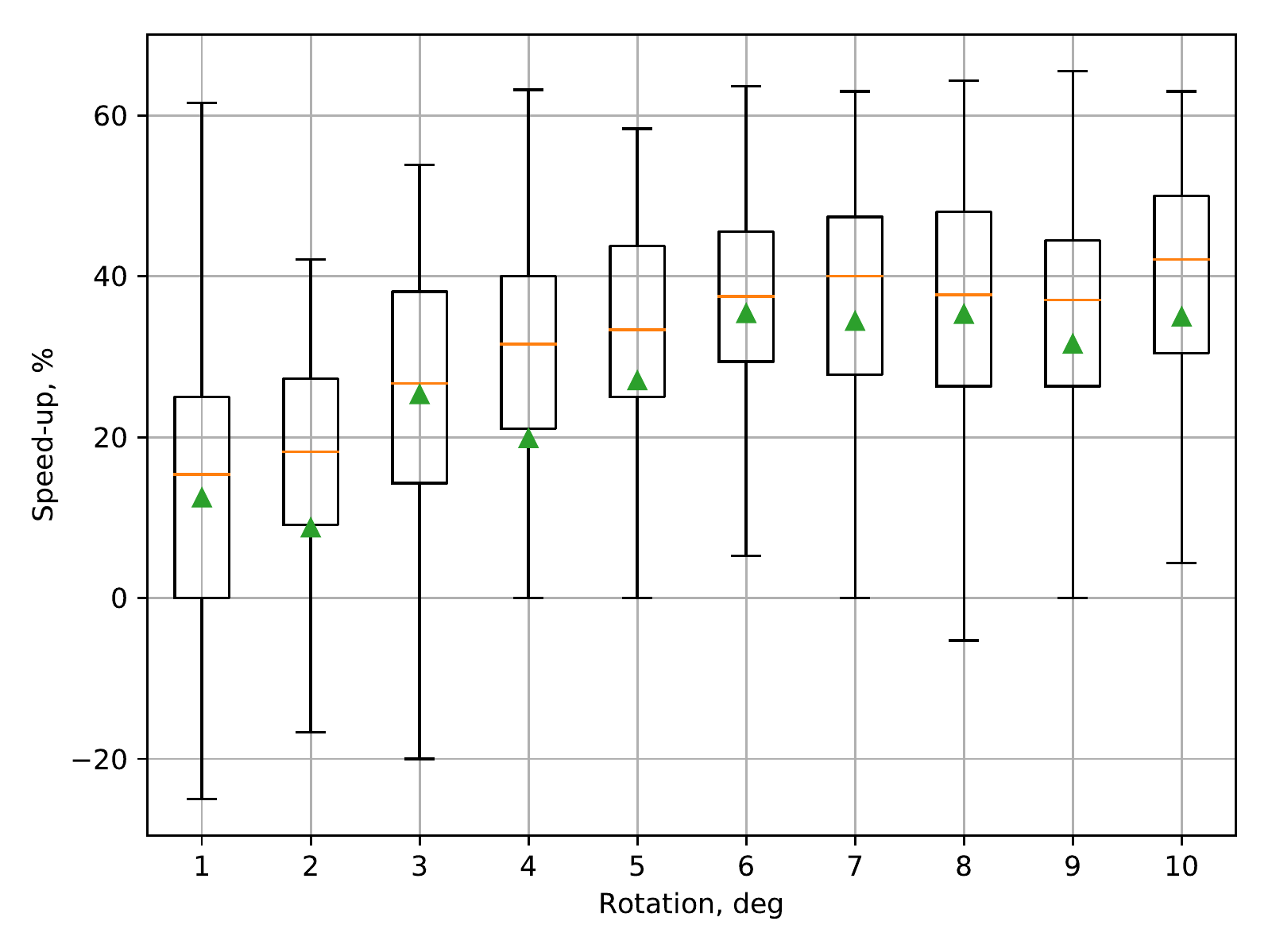}
    \caption{Box-plot of relative AA-ICP speed-up depending on random rotation angle for partial bunny scans.}
    \label{fig:bunny_angles}
\end{figure}

\begin{figure}
    \includegraphics[width=\columnwidth]{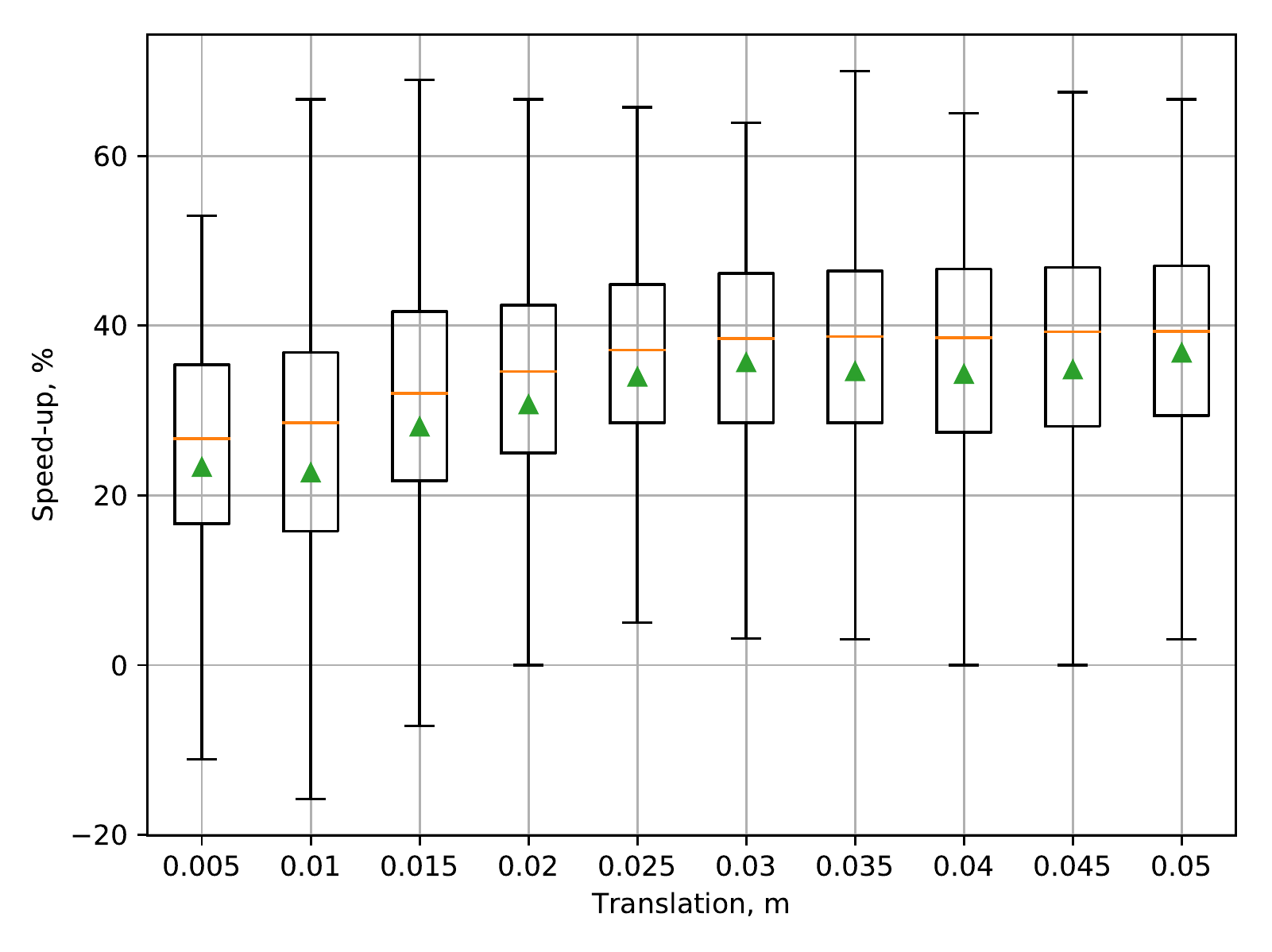}
    \caption{Box-plot of relative AA-ICP speed-up depending on random translation distance for partial bunny scans.}
    \label{fig:bunny_trans}
\end{figure}

In this work we have used two scans taken under 0 and 45 degrees (each contains approximately 40000 points). First we aligned them, and them applied random translations and rotations and measured acceleration of the AA-ICP compared to unmodified ICP. The same parameters have been used as in the previous section: $\varepsilon = 0.001$ and $\alpha_l = 10$.

To test acceleration properties of AA-ICP we introduced 1000 random iterations per given degree and ecorded relative acceleration compared to simple ICP. Results of this experiment can be seen in fig. \ref{fig:bunny_angles}. As we can see relative speed-up of AA-ICP rises up to 5 degrees and than stays on the slope of approximately 35\%.

Next experiment was done for translations, we did the same 1000 random translations for a given distance. Results can be seen in fig. \ref{fig:bunny_trans}, they look quite similar to experiment with random rotations. Here relative speed-up rises up to 2.5 cm and then stays on the slope of  approximately 35\%.

This behaviour can be explained by the fact that with bigger misalignment more iterations are needed to perform scan-matching, thus more room for AA-ICP to show its acceleration property, otherwise with small number of iterations (be it due to small initial misalignment or relaxed convergence criteria) its behaviour will be more similar to the simple ICP.

\section{Conclusions}

In this work we proposed and analyzed AA-ICP -- the novel modification of Iterative Closest Point algorithm based on Anderson acceleration. This method can be easily applied to existing implementations and has negligible runtime cost. It substantially reduces the number of iterations required for achieving desired scan-matching quality compared to the unmodified version.

We also implemented AA-ICP as part of Point Cloud Library and benchmarked it against unmodified ICP on real-world data, acquired by RGB-D camera and laser scanner. In addition to the successful demonstration of the acceleration properties of AA-ICP, we showed that with the same convergence criteria our algorithm provides better final results. Thus for same quality of convergence it can be used with less strict convergence criteria.

Further work can be done on the following topics: improving heuristics; benchmarking on other datasets; applying method to 2D case; and development of better theoretical foundation of Anderson acceleration for roto-translation group with not-fully contractive mappings. 

\addtolength{\textheight}{-12cm}   

\section*{Acknowledgment}
The research was supported by RSF (project No. 17-11-01376).

The authors are also grateful to Dr Evgeny Mironov from Center for Design, Manufacturing and Materials, Skolkovo Institute of Science and Technology for valuable suggestions and discussions.
\bibliographystyle{unsrt}
\bibliography{references}
\end{document}